\documentclass[final,12pt]{conf2025} 
\usepackage{algorithm}
\usepackage{algorithmic}
\usepackage{bm}
\usepackage{amsmath,amssymb}
\usepackage{wrapfig}
\newtheorem{thm}{Theorem}

\newtheorem{myDef}{Definition}
\newtheorem{assum}{Assumption}

\title[Global Convergence of Continual Learning on Non-IID Data]{Global Convergence of Continual Learning on Non-IID Data}
\usepackage{times}

\coltauthor{%
 \Name{Fei Zhu$^*$} \Email{zhfei2018@gmail.com}\\
 \addr Centre for Artificial Intelligence and Robotics, Hong Kong Institute of Science \& Innovation, CAS
 \AND
 \Name{Yujing Liu$^*$} \Email{yujing@amss.ac.cn}\\
 \addr Key Laboratory of Systems and Control, Academy of Mathematics and Systems Science, CAS
 \AND
 \Name{Wenzhuo Liu} \Email{liuwenzhuo2020@ia.ac.cn}\\
 \addr State Key Laboratory of Multimodal Artificial Intelligence Systems, Institute of Automation, CAS\\
 School of Artificial Intelligence, University of Chinese Academy of Sciences
 \AND
 \Name{Zhaoxiang Zhang$^\dagger$} \Email{zhaoxiang.zhang@ia.ac.cn}\\
 \addr State Key Laboratory of Multimodal Artificial Intelligence Systems, Institute of Automation, CAS\\
 School of Artificial Intelligence, University of Chinese Academy of Sciences\\\\
 $^*$ These authors contributed equally to this work\\
 $^\dagger$ Corresponding author: zhaoxiang.zhang@ia.ac.cn
}

\begin{document}

\maketitle

\begin{abstract}
Continual learning, which aims to learn multiple tasks sequentially, has gained extensive attention. However, most existing work focuses on empirical studies, and the theoretical aspect remains under-explored. Recently, a few investigations have considered the theory of continual learning only for linear regressions, establishes the results based on the strict independent and identically distributed (i.i.d.) assumption and the persistent excitation on the feature data that may be difficult to verify or guarantee in practice. To overcome this fundamental limitation, in this paper, we provide a general and comprehensive theoretical analysis for continual learning of regression models. 
By utilizing the stochastic Lyapunov function and martingale estimation techniques, we establish the almost sure convergence results of continual learning under a general data condition for the first time. 
Additionally, without any excitation condition imposed on the data, the convergence rates for the forgetting and regret metrics are provided.

\end{abstract}


\section{Introduction}
Continual learning (CL) \citep{ThrunRAS1995}, also known as incremental learning or lifelong learning, is an essential machine learning paradigm that enables the model to adapt and learn from a continuous stream of tasks over time \citep{van2022three, dohare2024loss}. Unlike traditional single-task learning, where a model is trained on a fixed dataset, CL emphasizes the sequential acquisition of knowledge across multiple tasks, requiring the model to retain and leverage previously learned information when learning the new task. For example, an autonomous car that is trained in a city would encounter other environments (e.g., forest, desert, countryside) with different data distributions \citep{verwimp2023clad}, and the car is supposed to adapt to those new environments while maintaining the ability in past environments. Besides, CL is also essential for many other applications such as foundational models \citep{yang2024recent, bommasani2021opportunities}, embodied robot \citep{lesort2020continual}, medical diagnosis \citep{lee2020clinical, perkonigg2021dynamic} and personalized recommendations \citep{xie2020kraken, mi2020ader}, in which the model must adapt to evolving data distributions over time without forgetting previous knowledge.
However, without storing old data, the model would lose previous knowledge when adapting to new data, which is known as catastrophic forgetting \citep{mccloskey1989catastrophic, goodfellow2013empirical}. As a result, CL becomes fundamentally more complex than single-task learning, requiring sophisticated mechanisms to balance the retention of older knowledge with the assimilation of new information.

\paragraph{Related Works.} In the past years, continual learning has received much attention in machine learning community, and a variety of empirical methods have been proposed \citep{de2021continual, van2022three}. In general, they can be divided into three main families. Regularization-based methods \citep{kirkpatrick2017overcoming, zenke2017continual, aljundi2018memory, li2017learning} identify and penalize the changes of important parameters of the original network when learning new tasks. Replay-based methods alleviate catastrophic forgetting by storing and replaying some raw data \citep{rebuffi2017icarl, chaudhry2018efficient, riemerlearning, jin2021gradient} or prototypes \citep{zhu2021prototype} of old tasks. Parameter-isolation methods \citep{rusu2016progressive, mallya2018packnet, serra2018overcoming, cortes2017adanet, xu2018reinforced} freeze the old network and dynamically extend it during the course of CL. More recent works build on pre-trained vision-language models \citep{wang2022learning}, large language models \citep{kecontinual}, or large multimodal models \citep{zeng2024modalprompt} and train small learnable parameters, e.g., prompt or LoRA \citep{hulora}, to dynamically instruct models in tackling tasks sequentially. Besides, some studies focus on more sophisticated CL settings such as long-tailed \citep{liu2022long}, few-shot \citep{mazumder2021few, tian2024survey} and federated \citep{yoon2021federated, guo2024pilora} continual learning.

Although significant progress has been made in the empirical aspect, theoretical
studies of continual learning are still largely unexplored.
In this context, the generalization error and forgetting for orthogonal gradient descent method \citep{zeng2019continual, farajtabar2020orthogonal} has been studied within the neural tangent kernel framework \citep{bennani2020generalisation, doan2021theoretical}. The impact of task similarity on forgetting has been investigated in teacher-student setup \citep{lee2021continual, asanuma2021statistical} and overparameterized regimes \citep{lin2023theory, evron2022catastrophic, dingunderstanding, goldfarbjoint}. Besides, \cite{kim2022theoretical} demonstrated the connection between task and class continual learning, and further proved that class continual learning is learnable \citep{kim2023learnability}. \cite{peng2023ideal} proposed an ideal continual learner and connected it with existing methods. \cite{evron2023continual} studied CL on a sequence
of separable linear classification tasks with binary labels, and developed upper bounds on the forgetting.
Due to the clear formulation and advantages in real-world applications (i.e. without storing old data or expanding the model), regularization-based continual learning has become a focus of recent theoretical studies \citep{li2023fixed, lin2023theory, peng2023ideal, dingunderstanding, zhao2024a}.

\paragraph{Limitations of Existing Work.} Despite these advancements, existing theoretical investigations on the continual
learning problem suffers from some fundamental limitations. 

Firstly, all prior theoretical works adopt a
naive locally-iid assumption, where the input data in each task is required to be independent and identically distributed (i.i.d.) to reduce the difficulty of theoretical analysis. For example, \cite{lee2021continual}, \cite{asanuma2021statistical}, \cite{swartworth2023nearly}, \cite{lin2023theory} and \cite{banayeeanzade2024theoretical} assumed all data are sampled from a standard Gaussian distribution. \cite{dingunderstanding} attempted to relax the standard Gaussian data assumption to the condition that population data covariance matrix satisfies the general fourth moment. However, the i.i.d. data assumption is still required in \citep{dingunderstanding} others studies \citep{li2023fixed, zhao2024a, peng2023ideal}.
Nevertheless, the i.i.d. data assumption often fails to align with real-world applications exhibiting more dynamic and complex distributions \citep{kejriwal2024challenges}. Recently, a few empirical studies call for establishing CL benchmarks in which distribution shifts arise gradually from the
passage of time \citep{lin2021clear, yao2022wild}.
Theoretically, it still remains an open question whether the i.i.d data assumption can be essentially removed for the parameter estimation of continual learning systems where data from multiple sources or environments emerge sequentially. 

Secondly, previous theoretical studies require other relatively
stringent assumptions that may not be applicable in practice. For example, \cite{zhao2024a} assumed that the accumulated covariance matrix is full-rank in the process of continual learning, which is known as the persistent excitation \citep{narendra1987persistent} that ensures the features are sufficiently diverse. Regard to system noise, \cite{evron2022catastrophic, evron2023continual} and \cite{peng2023ideal} assumed the model is noiseless and others \citep{lin2023theory, evron2022catastrophic, dingunderstanding, li2023fixed, zhao2024a, banayeeanzade2024theoretical} assumed the noise to be well-specific with a standard Gaussian distribution. However, real-world data often suffers from heavy-tailed or skewed noise due to certain physical processes or sensor errors.
Therefore, the persistent excitation and noise conditions are difficult to be satisfied or verified in many scenarios, especially in stochastic uncertain systems.

Thirdly, in terms of the model, previous investigations \citep{evron2022catastrophic, dingunderstanding, li2023fixed, peng2023ideal, goldfarbjoint, zhao2024a} mainly focus on linear regression and assume the existence of a share the same global minimizer, i.e., all tasks are generated by a linear model with the same regression coefficient. In practice, there might only exist an approximate common global minimizer due to distribution shifts, and the model is often nonlinear (e.g., with sigmoid \citep{hastie2009elements} or ReLU \citep{glorot2011deep} activations). Besides, some literature \citep{evron2022catastrophic, lin2023theory, dingunderstanding, banayeeanzade2024theoretical, goldfarbjoint} necessitates an over-parameterized regime that the data dimension is larger than the data size, which could be inapplicable for large-scale datasets.

\paragraph{Contributions.} The above limitations indicate that the theoretical investigation of continual learning is still in its infancy, and further effort is necessary to establish a deep understanding of the underlying principles. In this paper, we aim to establish the convergence of CL with general nonlinear regressions without i.i.d data assumptions, showing that the above-mentioned limitations can be largely relaxed or removed. 
The main contributions of this paper can be summarized as follows:
\begin{itemize}
\item We propose two novel continual learning algorithms to estimate the unknown parameters of the stochastic regression model. The former is applied to a class of nonlinear regression models with a shared global minimizer existing, while the latter can be applied to general linear regressions with only an approximate common global minimizer existing.

\item In the presence of a shared global minimizer, for the first time, we establish almost sure global convergence for nonlinear regressions under a general data condition based on stochastic Lyapunov function and martingale estimation techniques. Based on the parameter estimates, we provide the convergence rate for both forgetting and regret metrics without any excitation condition on the data.

\item When only an approximate common global minimizer exists, we establish the almost sure global convergence of the continual learning with general linear regressions under a general data condition, and provide the convergence rates for the forgetting and regret metrics.

\end{itemize}

The remainder of this paper is organized as follows: Section 2 presents the problem formulation, including the basic notation, model description and continual learning algorithm. Section 3 states the main results and the paper is concluded with some remarks in Section 4. The corresponding proofs of the main results are provided in the Appendix.

\section{Problem Formulation}
\paragraph{Basic Notations.}
For a $d$-dimensional column vector
$v\in{\mathbb{R}^d}$, ${v^{\top}}$ and $\|v\|$ denote its transpose and Euclidean norm, respectively.
Also, for a $d\times d$-dimensional matrix $A$, $\|A\|$ denotes its Euclidean norm (i.e., $(\lambda_{\max}\{AA^{\top}\})^{\frac{1}{2}}$), $\text{Tr}(A)$ is the trace, and the maximum and minimum eigenvalues are denoted by $\lambda_{\max}\{A\}$ and $\lambda_{\min}\{A\}$, respectively. 
For two matrices $A\in\mathbb{R}^{d\times d}$ and $B\in\mathbb{R}^{d\times d}$, $A>(\geq)B$ means that $A-B$ is a positive (semi-positive)-definite matrix. Throughout this paper, $\vert\cdot\vert$ denotes the determinant of the corresponding matrix. For a matrix sequence $\{A_k, k\geq 0\}$ and a positive scalar sequence $\{b_k, k\geq 0\}$, we say that $A_k= O(b_k)$ if there exists a constant $C>0$ such that $\left\|A_k\right\|\leq Cb_k$ holds for all $ k \geq 0$, and $A_k= o(b_k)$ if $\mathop {\lim }\limits_{k \to \infty } \|A_k\|/b_k=0$.

We use $\mathbb{E}[\cdot]$ to denote the mathematical expectation operator, and $\mathbb{E}[\cdot|\mathcal{F}_{k}]$ to denote the conditional mathematical expectation operator, where $\{\mathcal{F}_{k}\}$ is a sequence of non-decreasing $\sigma$-algebras. A sequence of random variable $\{\bm{x}_k\}$ is said to be adapted to $\{\mathcal{F}_{k}\}$ if $\bm{x}_k$ is $\mathcal{F}_{k}-$measurable for all $k \geq 0$. Furthermore, if $\mathbb{E}[\bm{x}_{k+1}|\mathcal{F}_{k}]=0$ for all $k \geq 0$, then the adapted sequence $\{\bm{x}_k, \mathcal{F}_{k}\}$ is called a martingale difference sequence.
For the linear space $\mathbb{R}^{d}~ (d\geq1)$, the weighted norm $\|\cdot \|_{\mathbf{Q}}$ associated with a positive definite matrix $\mathbf{Q}$ is defined as$\|\bm{x}\|_{\mathbf{Q}}=\bm{x}^\top\mathbf{Q}\bm{x}, \quad \forall \bm{x} \in \mathbb{R}^{d}.$

\begin{figure}[t]
    \begin{center}	
    \vskip -0.03 in 
\centerline{\includegraphics[width=0.67\columnwidth]{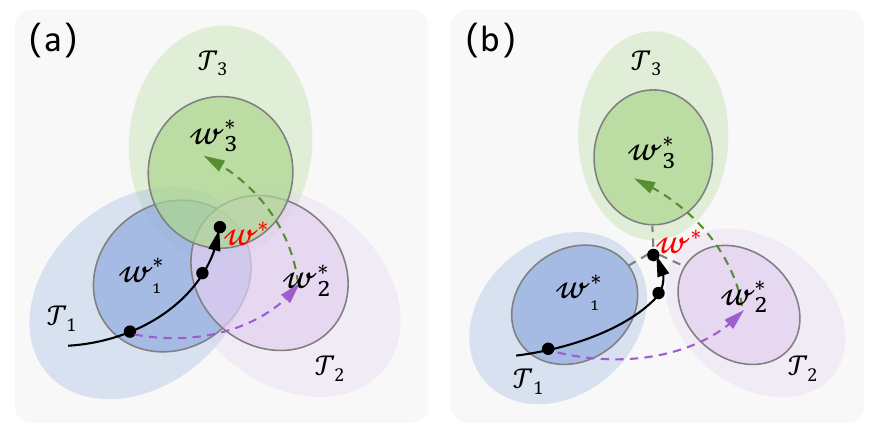}}
			\vskip -0.2 in
			\caption{Illustration of continual learning scenarios investigated in this paper. (a) All tasks share the same global minimizer $\bm{w}^*$. (b) A more general setting that relaxes case 1 into the existence of an approximate common global minimizer. Solid line with an arrow denotes the optimization trajectory of CL algorithm. Dashed line denotes naive sequent fine-tuning, which suffers from catastrophic forgetting.}
			\label{figure-1}
		\end{center}
		\vskip -0.25 in
\end{figure}

\paragraph{General Model Description.}
This paper focuses on the problem of learning a set of regression tasks $\mathcal{T} = \{0,...,t,...,m\}$ arriving sequentially in time. We assume data of the task at continual stage $t$ is generated by a standard stochastic nonlinear regression model:  
\begin{equation}\label{modelx}
    y_{t,i} = f(\bm{x}_{t,i}^\top \bm{w}_t^*, z_{t,i}), ~~~t\geq 0, i\geq 0,
\end{equation}  
where $f: \mathbb{R}\to\mathbb{R}$ is a known function, $\bm{w}_t^* \in \mathbb{R}^d$ is the unknown true parameter to be estimated, $\bm{x}_{t,i}\in \mathbb{R}^d$, $y_{t,i}\in \mathbb{R}$ and $z_{t,i}\in \mathbb{R}$ are the feature vector, model output, and the system noises, respectively. 
The data for each task $t \in \mathcal{T}$ is denoted as $\mathcal{D}_t = \{(\bm{x}^i_t, y^i_t)\}_{i=1}^{n_t}$, where $n_t$ is the sample size. 
At learning stage $t$, task $\mathcal{T}_t$ can be solved by minimizing some objective of the form
\begin{equation}  \label{opt_obj}
	\mathcal{S}_t:=\mathop{\arg\min}_{\bm{w} \in \mathcal{W}} \mathcal{L}_t(\bm{w};\mathcal{D}_t),
\end{equation}
where $\mathcal{S}_t$ is the set of global minimizers of some form of learning objective $\mathcal{L}_t$.
After learning task $t$, we get a new estimation $\bm{w}_t \in \mathcal{S}_t$. The aim of CL is to solve the tasks sequentially and find some ground truth in the hypothesis space $\mathcal{W}$ that minimizes the loss of all tasks. Specifically, we define the parameter variation process $\{\Delta_t\}$ as $ 
    \Delta_t = \bm{w}^*_t - \bm{w}^*,$
and investigate two cases:
\begin{itemize}
\item \textbf{Case 1:} $\cap_{t=1}^{m} \mathcal{S}_t \neq \varnothing$, which assumes that all tasks are generated by the same model $\bm{w}^*$, sharing a common global minimizer, as illustrated in Figure \ref{figure-1} (a). If an algorithm can find the common global minimizer, we say it never forgets.
\item \textbf{Case 2:} $\cap_{t=1}^{m} \mathcal{S}_t = \varnothing$ while $\frac{1}{m+1}\sum^{m}_{t=1}\Delta_t =0$, which is a more general setting that relaxes the Case 1 into the existence of approximate common global minimizer (Figure \ref{figure-1} (b)), i.e., the optimal model parameters might be different for different tasks, and continual learning without forgetting is infeasible.
\end{itemize}

Case 1 assumes that all tasks share a common global minimizer, i.e., $\bm{w}^* \in \cap_{t=1}^{m} \mathcal{S}_t$. This is actually a common setting in existing CL studies \citep{evron2022catastrophic, dingunderstanding, banayeeanzade2024theoretical, goldfarbjoint, peng2023ideal, zhao2024a, li2023fixed}. However, those works only investigated this setting with linear regressions. In this paper, we extend it to general nonlinear setting including both nonlinear regression and classification. In addition, we further study a general case 2 where $\cap_{t=1}^{m} \mathcal{S}_t = \varnothing$ and only an approximate common global minimizer $\bm{w}^* \in \mathcal{W}$ exists for linear models, as illustrated in Figure \ref{figure-1}.

\paragraph{Performance Metric.} The main goal in continual learning is to adapt to new tasks while not forgetting the knowledge of previous tasks. For example, an autonomous car should maintain the ability in previous environments after being trained in a new environment. Besides, we also expect that with knowledge accumulation during continual learning, the model is more robust on tasks with unseen domains, e.g., the autonomous car should be robust when encountering unseen environments before learning it. To characterize the above two aspects, we formally define the forgetting and regret as follows  \citep{evron2022catastrophic,doan2021theoretical}:
\begin{myDef}\label{def2} (Forgetting \& Regret) Let $\mathcal{D}_k = \{(\bm{x}^i_k, y^i_k)\}_{i=1}^{n_k}$ be the data of task $k \in \{1,2,...,t\}$. The forgetting at continual stage $t$ is the average loss of already seen tasks, and the regret is the average loss on unseen tasks, i.e.,
\[
F_{t} := \frac{1}{t} \sum_{k=1}^{t} \mathcal{L}(\bm{w}_t; \mathcal{D}_k), ~~~~R_{t} := \frac{1}{t} \sum_{k=1}^{t} \mathcal{L}(\bm{w}_{k-1}; \mathcal{D}_{k}).
\]
\end{myDef}

\subsection{Case 1: Continual Learning with Common Global Minimizer}\label{model}

Before presenting the continual learning algorithm for the case with a common global minimizer, we introduce the following assumptions:

\begin{assum}\label{assum2}
For every task $t$, the number of samples $n_t$ is finite and the feature $\{x_{t,i}\}$ satisfies:
\begin{equation} \label{bound} 
\sup_{t\geq0}\sup_{i\geq0}\|\bm{x}_{t,i}\| <L, \quad \text{a.s.}
\end{equation}
\end{assum}
\begin{remark}
The assumption that $n_t$ is finite is meaningful for continual learning that learns a stream of tasks incrementally. For the case $n_t \to \infty$, our analysis is also applicable and similar results can be obtained.
The assumption that the norm of the data is bounded generalizes that in \citep{evron2022catastrophic}. We remark that this is a mild condition that is commonly used in learning theory literature \citep{novikoff1962convergence, vapnik2013nature, shalev2014understanding}. In practice, raw data is almost always subjected to normalization or standardization before being fed into models \citep{nixon2019feature, camacho2018word}. Besides, data generation processes in real-world scenarios inherently follow physical boundaries. Therefore, the above boundedness assumption can be easily achieved in practice.
\end{remark}

To estimate the true parameter $\bm{w}^*$, we design regularized continual learning algorithms with learning objective at the continual stage $t$ is given by
\begin{equation}\label{objjj}
\begin{aligned}
\mathcal{L}_t(\bm{w}) = \sum_{i=1}^{n_t}\mathcal{L}\left(\bm{x}_{t,i}^{\top}\bm{w},y_{t,i}\right) + \frac{1}{2}\|\bm{w}-\bm{w}_{t-1}\|^2_{\mathbf{Q}_{t-1}}.
\end{aligned}
\end{equation}
Define $g_1(x,y) = \frac{\partial \mathcal{L}(x, y)}{\partial x}$ and $g_2(x,y) = \frac{\partial^2 \mathcal{L}(x, y)}{\partial^2 x}$.
For our theoretical analysis, we need the following mild assumption on the objective function:
\begin{assum}\label{assum1}
The objective function satisfies the following properties:

i) For every sample $\bm{x}_{t,i}$ in every task $t$, we have 
$$\mathbb{E} \left[ g_1\left(\bm{x}_{t,i}^{\top}\bm{w}^{*}, y_{t,i}\right) | \mathcal{F}_{t} \right] = 0;$$  

ii) For any $C\geq0$ and any $|\xi_{t,i}|\leq C$, we have
$$
\begin{aligned}
\inf_{t\geq0} \inf_{|\xi_{t,i}|\leq C}\mathbb{E}\left[g_2(\xi_{t,i}, y_{t,i})|\mathcal{F}_t\right]\geq \underline \mu>0,\ \text{and}\
\sup_{|\xi_{t,i}|\leq C}g_2(\xi_{t,i}, y_{t,i})\leq \bar \mu <\infty, \ \hbox{a.s.}
\end{aligned}
$$
\end{assum}

\begin{remark}\label{rm2}
This assumption covers a wide range of commonly used regression and classification models, which are detailed below:

1) \textbf{Standard linear regression model} widely used for the data fitting tasks \citep{zhao2024a, li2023fixed, evron2022catastrophic, peng2023ideal, lin2023theory}:
$$y_{t,i} = \bm{x}_{t,i}^{\top}\bm{w}^{*} + z_{t,i},$$
where the system noise $\{z_{t,i}, \mathcal{F}_t\}$ is a martingale difference sequence.
Consider the following mean square loss function
$$
\begin{aligned}
\mathcal{L}(\bm{x}_{t,i}^{\top}\bm{w}, y_{t,i})=\frac{1}{2}(\bm{x}_{t,i}^{\top} \bm{w} - y_{t,i})^2.
\end{aligned}
$$
In this case, $g_1(\bm{x}_{t,i}^{\top} \bm{w}^{*},y_{t,i}) =\bm{x}_{t,i}^{\top} \bm{w}^{*} - y_{t,i}=-z_{t,i}$ and $
g_2(\bm{x}_{t,i}^{\top} \bm{w}^{*} - y_{t,i}) =1$, which clearly satisfy our assumption. 

2) \textbf{Logistic regression model} widely used for the classification tasks \citep{tortajada2015incremental,evron2023continual, de2021continual, mai2022online}:
$$y_{t,i}=\frac{1}{1+\exp(-\bm{x}_{t,i}^{\top}\bm{w}^{*})}+z_{t,i},$$
where $z_{t,i}$ is the same as in the previous example.
Consider the following cross-entropy loss function
$$
\begin{aligned}
\mathcal{L}(\bm{x}_{t,i}^{\top}\bm{w}, y_{t,i})=-y_{t,i}\log(f(\bm{x}_{t,i}^{\top}\bm{w}))-(1-y_{t,i})\log(1-f(\bm{x}_{t,i}^{\top}\bm{w})).
\end{aligned}
$$
where $f(x) = 1/(1+\exp(-x))$.
Then it follows that 
$$g_1(\bm{x}_{t,i}^{\top} \bm{w}^{*},y_{t,i})=f(\bm{x}_{t,i}^{\top} \bm{w}^{*})-y_{t,i}=-z_{t,i}\ \ \text{and} \ \ g_2(\bm{x}_{t,i}^{\top} \bm{w}^{*},y_{t,i})=f(\bm{x}_{t,i}^{\top} \bm{w}^{*})(1 - f(\bm{x}_{t,i}^{\top} \bm{w}^{*})),$$ which also satisfy our assumption.

\begin{algorithm}[t]
\caption{Case 1: Continual Learning with Common Global Minimizer}
\label{alg1}
\begin{algorithmic}
\STATE {\bfseries Initialization:} $\bm{w}_0$, $\mathbf{Q}_{0}=I$
\STATE {\bfseries Iterative update for each task $t\in \mathcal{T}$:}
\begin{subequations}\label{algg1}
\begin{align}
&\bm{w}_{t}=\Pi_{\mathbf{Q}_{t}}\left[\bm{w}_{t-1}-\mathbf{Q}_{t}^{-1}\left(\sum_{i=1}^{n_t} g_1(\bm{x}_{t,i}^{\top} \bm{w}_{t-1},y_{t,i})\bm{x}_{t,i}\right)\right],\label{alg1a}\\
&\mathbf{Q}_{t} = \mathbf{Q}_{t-1} + \mu^2\sum_{i=1}^{n_{t}} \bm{x}_{t, i} \bm{x}_{t,i}^{\top},\label{alg1b}
\end{align}
\end{subequations}
where $\mu^2\in(0,\underline \mu)$ is the adaptation gain scalar sequence, and the projection operator $\Pi_{\mathbf{Q}_{t+1}}(x)$ is defined in (\ref{projector}).
\end{algorithmic}
\end{algorithm}
3) \textbf{Stochastic saturation model} widely used for censored or truncated regression tasks \citep{breen1996regression,daskalakis2019computationally}:
$$
\begin{aligned}
&s_{t,i}=\bm{x}_{t,i}^{\top}\bm{w}^{*} + z_{t,i},\\
&y_{t,i}=\left\{
\begin{aligned}
& U,\ \ \text{if}\ \ s_{t,i}\geq u,\\
& s_{t,i},\ \ \text{if}\ \ l\leq s_{t,i}\leq u, \\
& L,\ \ \text{if}\ \ s_{t,i}\leq l,
\end{aligned}
\right.
\end{aligned}
$$
where the system noise $\{z_{t,i}, \mathcal{F}_t\}$ is dependent with a zero mean and variance $1$ Gaussian distribution. This includes the widely used models with ReLU activation \citep{nair2010rectified, glorot2011deep}.
Consider the following negative log-likelihood loss function:
\begin{equation*}
\begin{aligned}
&\mathcal{L}(\bm{x}_{t,i}^{\top}\bm{w}, y_{t,i})=-\log \mathcal{Q}_{n}(\bm{x}_{t,i}^{\top}\bm{w}, y_{t,i})\\
=&-\delta_{t,i}\log(F(l-\bm{x}_{t,i}^{\top}\bm{w}))-(1-\delta_{t,i}-\bar{\delta}_{t,i})\log (f(y_{t,i}-\bm{x}_{t,i}^{\top}\bm{w}))-\bar{\delta}_{t,i}\log(1-F(u-\bm{x}_{t,i}^{\top}\bm{w})),
\end{aligned}
\end{equation*}
where $\delta_{t,i}=I_{\{y_{t,i}=L\}}$, $\bar{\delta}_{t,i}=I_{\{y_{t,i}=U\}}$, $F(x)$ and $f(x)$ are the cumulative distribution function and probability dense function of normal distribution, respectively.
In this case, we have
$$
\begin{aligned}
&g_1(\bm{x}_{t,i}^{\top} \bm{w}^{*},y_{t,i})=\delta_{t,i}\frac{f(l-\bm{x}_{t,i}^{\top} \bm{w}^{*})}{F(l-\bm{x}_{t,i}^{\top} \bm{w}^{*})}+(1-\delta_{t,i}-\bar \delta_{t,i})\frac{f'(y_{t,i}-\bm{x}_{t,i}^{\top} \bm{w}^{*})}{f(y_{t,i}-\bm{x}_{t,i}^{\top} \bm{w}^{*})}-\bar \delta_{t,i}\frac{f(u-\bm{x}_{t,i}^{\top} \bm{w}^{*})}{1-F(u-\bm{x}_{t,i}^{\top} \bm{w}^{*})},\\
&g_2(\bm{x}_{t,i}^{\top} \bm{w}^{*},y_{t,i})=
\frac{(l-\bm{x}_{t,i}^{\top} \bm{w}^{*})f(l-\bm{x}_{t,i}^{\top} \bm{w}^{*})F(l-\bm{x}_{t,i}^{\top} \bm{w}^{*})+f^2(l-\bm{x}_{t,i}^{\top} \bm{w}^{*})}{F^{2}(l-\bm{x}_{t,i}^{\top} \bm{w}^{*})}\delta_{t,i}\\
&\ \ \ \ \ \ \ \ \ \ \ \ +\frac{(\bm{x}_{t,i}^{\top} \bm{w}^{*}-u)f(\bm{x}_{t,i}^{\top} \bm{w}^{*}-u)F(\bm{x}_{t,i}^{\top} \bm{w}^{*}-u)+f^2(\bm{x}_{t,i}^{\top} \bm{w}^{*}-u)}{F^{2}(\bm{x}_{t,i}^{\top} \bm{w}^{*}-u)}\bar\delta_{t,i}+(1-\delta_{t,i}-\bar{\delta}_{t,i}).
\end{aligned}
$$
It is easy to verify that $\mathbb{E}\left[ g_1\left(\bm{x}_{t,i}^{\top}\bm{w}^{*}, y_{t,i}\right) | \mathcal{F}_{t}\right]=0$.
Besides, by the facts that $\frac{xf(x)F(x)+f^2(x)}{F^2(x)}\in (0,1)$ for $x\in(-\infty,+\infty)$ and it is a strictly decreasing function (see Lemma 1 in \citep{zhao2016iterative}), there exists a positive $\underline\mu$ such that $g_2(\bm{x}_{t,i}^{\top} \bm{w}^{*},y_{t,i})\geq \underline \mu$ for $|\bm{x}_{t,i}^{\top} \bm{w}^{*}|$ in any bounded set. So our assumption is still satisfied.
\end{remark}

We propose a continual learning algorithm (i.e., Algorithm \ref{alg1}) to estimate the parameter $\bm{w}$ recursively. 
In Algorithm \ref{alg1}, we introduce a projection operator to ensure the boundedness of the parameter estimates:
\begin{equation}\label{projector}
\Pi_{\mathbf{Q}_{t}}(x)=\mathop{\arg\min}_{w\in \mathcal{H}} \|x-w\|_{\mathbf{Q}_{t}},
\end{equation}
here $\mathcal{H}=\{w: \|w\|\leq M\}$ with $\bm w^*\in \mathcal{H}$.
Furthermore, we design an adaptation gain scalar $\mu$ in order to ensure the convergence of parameter estimates by leveraging the lower bound of the second-order derivative for the loss function. 
Specifically, for the learning tasks of linear regression models, the scalar factor $\mu$ can be set to $1$, which is consistent with the classical least squares algorithm.

\subsection{Case 2: Continual Learning without Common Global Minimizer}\label{model2}
When no global minimizer is shared among tasks, a few recent works studied this setting \citep{lin2023theory, banayeeanzade2024theoretical, goldfarbjoint}. However, they require a stringent assumption that the linear regression model being over-parameterized, which may be invalid when involving large datasets. Moreover, the bounds of forgetting in those work are yields in an expectation form with i.i.d data assumption and Gaussian noise condition. Differently, we investigate general linear regressions where an approximate common global minimizer exist, and establish the almost sure global convergence results for the first time. Particularly, recent state-of-the-art CL methods \citep{zhuang2022acil, mcdonnell2024ranpac} typically build upon pre-trained backbone with a linear model, and consider that linear regression result can also be applied to complex models like deep networks in the neural kernel regime \citep{doan2021theoretical, evron2022catastrophic}, thus we leave the extension of our results in case 2 to the more complex nonlinear setting as future work.

To be specific, we consider the model (\ref{modelx}) with a linear function $f$, i.e., $y_{t,i} = \bm{x}_{t,i}^\top \bm{w}_t^*+ z_{t,i}, \ t\geq 0,\ i\geq 0$.
In order to establish a rigorous result, we need some specific assumptions on the true parameters $\{\bm{w}_t^*\}$, the system noise $\{z_{t,i}\}$ and the feature data $\{x_{t,i}\}$.
Before presenting these assumptions, we first define two $\sigma$-sequences $\mathcal{F}_t$ and $\mathcal{G}_t$ as: $\mathcal{F}_t=\sigma\{\bm{x}_{s,i}, z_{s-1,i}, \bm{w}_{s}^*, 1 \leq i \leq n_s, 0\leq s\leq t\}, $ and $\mathcal{G}_t=\sigma\{\bm{x}_{s,i}, z_{s-1,i}, \bm{w}_{s-1}^*, 1 \leq i \leq n_s, 0\leq s\leq t\}$.
\begin{assum}\label{assumpara}
The parameter variation $\{\bm{w}^*_t - \bm{w}^*,\mathcal{G}_{t}\}$ is a martingale difference sequence and there exists $\beta>2$ such that $\sup_{t\geq 0}\mathbb{E}\left[\|\bm{w}^*_t - \bm{w}^*\|^{\beta}|\mathcal{G}_{t}\right]<\infty, \ \hbox{a.s.}$
\end{assum}
\begin{remark}
We provide some examples that satisfy our assumption:

1) For each task \( t \), the true parameter is \( \bm{w}_t^* = \bm{w}^* + \Delta_t \), where $\Delta_t$ is a Gaussian random vector (i.e., \( \Delta_t \sim \mathcal{N}(0, \sigma^2 \bm{I}) \)) or be uniformly sampled from $[-c, c]^d$, and $\Delta_t$ is independent of all previous tasks.
    
2) For each task $t$, assume that the true parameter $\bm{w}_t^{*}$ is i.i.d. sampled from a parameter set $\{\bm{w}_1, \bm{w}_2,...,\bm{w}_s\}$ with a distribution $P(\bm{w}_t^{*}=\bm{w}_i)=\pi_i$ and $\sum\nolimits_{i=1}^s\pi_{i}=1$, as in the mixed linear regression learning problem \citep{liuconvergence}. Then the parameter variation process $\Delta_t = \bm{w}^*_t - \bm{w}^*$ satisfies Assumption \ref{assumpara} with $\bm{w}^* = \sum^{s}_{i=1}\pi_i \bm{w}_i$. 
\end{remark}

\begin{assum}\label{assumnoise}
The noise sequence $\{\sum_{i=1}^{n_t}z_{t,i}, \mathcal{F}_t\}$ is a martingale difference sequence and there exists a $\alpha> 2$ such that  $
\sup_{t\geq0} E[|\sum\nolimits_{i=1}^{n_t}z_{t,i}|^\alpha|\mathcal{F}_t] < \infty, \ \text{a.s.}$
\end{assum}

\begin{assum}\label{assum2case2}
For every task $t$, the number of samples $n_t$ is finite. Besides, there exists a constant $\delta \in [0,\frac{1}{2})$ such that $\sum\nolimits_{i=1}^{n_t}\|\bm{x}_{t,i}\|^2 = O(t^{\delta}), \quad \text{a.s.}$
\end{assum}
\begin{remark}
We remark that $\sum\nolimits_{i=1}^{n_t}\|\bm{x}_{t,i}\|^2=O(t^{\delta})$ is a mild condition that can be easily satisfied. Here we provide several widely used examples \citep{chen2}: 

1) If $\{\bm{x}_{t,i}\}$ is bounded as that in Assumption \ref{assum2} of Case 1, then the parameter $\delta$ can be zero.

2) If $\{\bm{x}_{t,i}\}$ is i.i.d. Gaussian as used in previous investigations \citep{zhao2024a, li2023fixed, lin2023theory}, then $\sum\nolimits_{i=1}^{n_t}\|\bm{x}_{t,i}\|^2=O(\log t)$ and the parameter $\delta$ can be chosen as any small positive constant; 

3) If there exists a constant $\alpha>4$ such that $\sup_{t\geq0}\mathbb{E}[\|\bm{x}_{t,i}\|^{\alpha}]<\infty,$ 
then the parameter $\delta$ can be any constant in $\left(\frac{2}{\alpha},\frac{1}{2}\right)$.
\end{remark}

The learning objective for CL of the general linear regression model at the stage $t$ is given by
\begin{equation}\label{obj2}
\begin{aligned}
\mathcal{L}_t(\bm{w})=\beta_t \sum\limits_{i=1}^{n_t} (\bm{x}_{t,i}^{\top}\bm{w} - y_{t,i})^2 + \|\bm{w}-\bm{w}_{t-1}\|^2_{\mathbf{Q}_{t-1}}.
\end{aligned}
\end{equation}
We propose a continual learning algorithm (i.e., Algorithm \ref{alg2}) to estimate the parameter recursively. Unlike directly using Algorithm \ref{alg1} for linear regression model with mean square loss, in Algorithm \ref{alg2}, we introduce an additional adaptation gain sequence $\{\beta_t\}$ to address the difficulty that caused by the possible unboundedness of the feature data $\{x_{t,i}\}$. Besides, by adjusting the value of $\{\beta_t\}$, we can allocate different importance weights for different tasks. For instance, in healthcare applications, our algorithm allows the model to allocate importance weights when learning from different task streams based on task significance and operational constraints \citep{feng2022clinical}.

\begin{algorithm}[t]
\caption{Case 2: Continual Learning without Common Global Minimizer}
\label{alg2}
\begin{algorithmic}
\STATE {\bfseries Initialization:} $\bm{w}_0$, $\mathbf{Q}_{0}=I$
\STATE {\bfseries Iterative update for each task $t\in \mathcal{T}$:}
\begin{subequations}\label{algg2}
\begin{align}
&\bm{w}_{t}=\bm{w}_{t-1}+ \mathbf{Q}_{t}^{-1}\beta_t\sum\limits_{i=1}^{n_t}\bm{x}_{t,i}(y_{t,i} -\bm{x}_{t,i}^{\top}\bm{w}_{t-1}),\label{alg2a}\\
&\mathbf{Q}_{t} = \mathbf{Q}_{t-1} + \beta_{t}\sum\limits_{i=1}^{n_t}\bm{x}_{t,i}\bm{x}_{t,i}^{\top},\label{alg2b}
\end{align}
\end{subequations}
where $\{\beta_t\}$ is an adaptation gain sequence.
\end{algorithmic}
\end{algorithm}

\section{Main Results} \label{main_results}
This section presents the main results for Case 1 \& 2, including the convergence results, the convergence rates of forgetting and regret metrics. Our analysis leverages stochastic Lyapunov functions and martingale estimation techniques, which help us to relax some commonly used yet stringent conditions such as the i.i.d data assumption, the persistent excitation and noise conditions of classical analysis in continual learning, making it highly applicable for future theoretical investigation. The detailed proof of those results can be found in the Appendix.

\subsection{Global Convergence Results for Case 1} \label{domain}
For case 1 where a shared global minimizer exists, we prove the convergence of the continual learning algorithm in a nonlinear setting which includes widely used models such as standard linear regression, logistic regression based classification, and the stochastic saturation model. Then based on the proposed algorithm, we provide the upper bounds of accumulated forgetting and regrets.

\begin{thm}\label{thm1}
Under Assumptions $\ref{assum2}$-$\ref{assum1}$, the estimation error generated by Algorithm $\ref{alg1}$ has the following upper bound as $m\to\infty$:
    \begin{equation}\label{rt}
        \begin{aligned}
            \left\|\widetilde{\bm{w}}_m \right\|^{2}=&O\left(\frac{\log m }{\lambda_{\min}(m)}\right), \;\;\mathrm{a.s}.,
        \end{aligned}
    \end{equation}
where $\widetilde{\bm{w}}_m = \bm{w}_m - \bm{w}^{*}$ and $\lambda_{\min}(m) = \lambda_{\min}\left\{\mathbf{Q}_0 +  \sum\limits_{t=1}^{m} \sum\limits_{i=0}^{n_t}  \bm{x}_{t,i} \bm{x}_{t,i}^\top\right\}.$
\end{thm}

\begin{remark}
Theorem \ref{thm1} shows that if
\begin{equation}\label{remarkccc}
\log m =o\left(\lambda_{\min}(m)\right),\;\; \mathrm{a.s}.
\end{equation}
as $m \rightarrow \infty$, then the estimate $\bm{w}_{m}$  will converge to the true unknown parameter $\bm{w}^{*}$ almost surely.
We remark that (\ref{remarkccc}) is a general data condition that does not impose strict informational richness on individual tasks, such as the i.i.d. Gaussian or the traditional persistent excitation i.e., \,  $m=O\left(\lambda_{\min}(m)\right), \ \hbox{a.s.}$
In fact, condition (\ref{remarkccc}) is exactly the \textbf{weakest possible} excitation condition established for the convergence of classical least squares of linear regression models \citep{lai1982least}.
Moreover, we also note that the convergence rate established in Theorem \ref{thm1} is essentially in terms of the increase of the number of observations rather than the number of iterations in computation, which makes it also applicable for online continual learning setting \citep{mai2022online}.
\end{remark}

To establish the performance analysis of our continual learning algorithm, we first introduce the following index that quantifies the optimal loss value over all past $t$ tasks at a shared global minimizer $\bm{w}^*$:
\begin{equation}
\mathcal{L}^*_t=\frac{1}{t}\sum\limits_{k=1}^t\sum_{i=1}^{n_k}\mathcal{L}\left(\bm{x}_{k,i}^{\top}\bm{w}^*,y_{k,i}\right),
\end{equation}
which is usually achievable by joint training (a standard approach where all tasks are trained simultaneously).
\begin{thm}\label{thm2}
Under Assumptions $\ref{assum2}$-$\ref{assum1}$, the forgetting metric defined in Definition \ref{def2} has the following upper bound:
\begin{equation}\label{forget1}
F_{t}=\frac{1}{t} \sum_{k=1}^{t} \mathcal{L}(\bm{w}_t; \mathcal{D}_k)=\frac{1}{t} \sum\limits_{k=1}^t\sum_{i=1}^{n_k}\mathcal{L}\left(\bm{x}_{k,i}^{\top}\bm{w}_t,y_{k,i}\right)=\mathcal{L}^*_t+O\left(\sqrt{\frac{\log t}{t}}\right), \ \hbox{a.s.}
\end{equation}
Moreover, the regret metric defined in Definition \ref{def2} has the following upper bound:
\begin{equation}\label{regret2}
R_{t}=\frac{1}{t} \sum_{k=1}^{t} \mathcal{L}(\bm{w}_{k-1}; \mathcal{D}_{k})=\frac{1}{t} \sum\limits_{k=1}^t\sum_{i=1}^{n_k}\mathcal{L}\left(\bm{x}_{k,i}^{\top}\bm{w}_{k-1},y_{k,i}\right)=\mathcal{L}^*_t+O\left(\frac{\log t}{t}\right), \ \hbox{a.s.}
\end{equation}
\end{thm}
\begin{remark} 
Actually, the term $\mathcal{L}^*_t$ can be viewed as a measure of system noise. As shown in the examples in Remark \ref{rm2}, the optimal loss value is usually determined by the variance of system noise. In noiseless settings, such as those in \citep{evron2022catastrophic, evron2023continual, peng2023ideal}, the optimal loss value is zero naturally. 
We note that the convergence result in an almost sure sense established in Theorem \ref{thm2} is better than the convergence in mean in existing studies \citep{lin2023theory}.
Moreover, we establish the convergence rate for the forgetting and regret metrics for the first time and the convergence rate of regret is potentially the \textbf{best possible} in the noise setting among all continual learners.
In addition, these convergence results do not require any excitation or richness condition on the feature data, and instead, they only require the boundedness of the feature data, which is considerably weaker than the persistent excitation condition used in the existing literature.
\end{remark}

\subsection{Global Convergence Results for Case 2} \label{domain2}
For case 2 where only an approximate common global minimizer exists, we establish the convergence of the continual learning with general linear regressions and provide the convergence rates of forgetting and accumulated regrets, respectively.

\begin{thm}\label{thm3}
Under Assumptions $\ref{assumpara}$-$\ref{assum2case2}$, the estimation error produced by Algorithm $\ref{alg2}$ with $\beta_t=\frac{1}{t^{\delta}}$ has the following upper bound:
\begin{equation}\label{rt33}
    \begin{aligned}
        \left\|\widetilde{\bm{w}}_m \right\|^{2}=&O\left(\frac{m^\delta \log m  }{\lambda_{\min}(m)}\right), \;\;\mathrm{a.s}.,
    \end{aligned}
\end{equation}
where $\widetilde{\bm{w}}_m = \bm{w}_m - \bm{w}^{*}$, $\lambda_{\min}(m)$ is defined in Theorem \ref{thm1}, and $\delta$ is defined in Assumption \ref{assum2case2}.
\end{thm}

\begin{remark}
Theorem \ref{thm1} shows that if as $\sum_{t=1}^{m} n_t \rightarrow \infty$, we have
\begin{equation}\label{cccc}
    m^\delta \log m=o\left(\lambda_{\min}(m)\right),\;\; \mathrm{a.s}.,
\end{equation}
then the estimate $\bm{w}_{m}$ will converge to the parameter $\bm{w}^{*}$ almost surely.
Moreover, when the feature data is bounded, the convergence rate can be the same as that in Theorem \ref{thm1}.
\end{remark}

\begin{figure}[t]
	\begin{center}	
		\centerline{\includegraphics[width=\columnwidth]{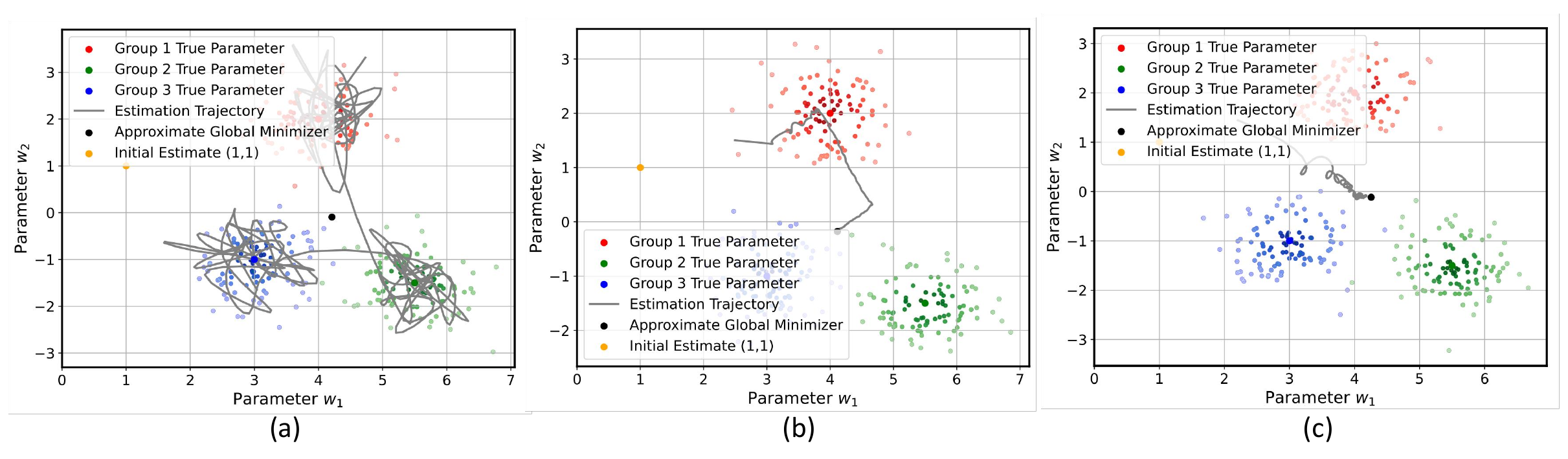}}
		\vskip -0.15 in
		\caption{Numerical demonstration of Case 2: (a) SGD suffers from catastrophic forgetting when continually learning 100 tasks. (b) Our algorithm can successfully find the approximate common global minimizer in sequential and (c) random learning orders.}
		\label{figure-2}
	\end{center}
	\vskip -0.15 in
\end{figure}
To further establish the performance analysis, we introduce the following two indexes:
\begin{equation}
\mathcal{L}^*_t=\frac{1}{t}\sum\limits_{k=1}^t\sum_{i=1}^{n_k}\left(y_{k,i}-\bm{x}_{k,i}^{\top}\bm{w}^*\right)^2,\ \ \ \mathcal{P}^*_t=\frac{1}{t}\sum\limits_{k=1}^t\sum_{i=1}^{n_k}\left(y_{k,i}-\bm{x}_{k,i}^{\top}\bm{w}_t^*\right)^2
\end{equation}
where $\mathcal{L}^*_t$ is the optimal loss value at a shared global minimizer $\bm{w}^*$ over all past $t$ tasks, and $\mathcal{P}^*_t$ is the optimal value of all past task $t$ with its true parameter $w_{t}^*$.
\begin{thm}\label{thm4}
Under Assumptions $\ref{assumpara}$-$\ref{assum2case2}$, the forgetting metric has the following upper bound:
\begin{equation}\label{forget3}
F_{t}=\frac{1}{t} \sum\limits_{k=1}^t\sum_{i=1}^{n_k}\left(y_{k,i}-\bm{x}_{k,i}^{\top}\bm{w}_t\right)^2=\mathcal{L}^*_t+O\left(\frac{t^{\delta}(\log t)^{1/2}}{t^{1/2}}\right), \ \hbox{a.s.}
\end{equation}
Moreover, the regret metric has the following upper bound:
\begin{equation}\label{regret4}
R_{t}=\frac{1}{t} \sum\limits_{k=1}^t\sum_{i=1}^{n_k}\left(y_{k,i}-\bm{x}_{k,i}^{\top}\bm{w}_{k-1}\right)^2=\mathcal{L}^*_t+O\left(\frac{t^{\delta}\log t}{t}\right), \ \hbox{a.s.}
\end{equation}
\end{thm}
\begin{remark}
We note that similar to Theorem \ref{thm2}, the almost sure convergence results for forgetting and regret metrics do not impose any excitation condition on the feature data, which is better than the convergence in mean under i.i.d. Gaussian data assumptions established in existing studies \citep{lin2023theory}. Moreover, unlike case 1 where the optimal value $\mathcal{L}^*_t$ is entirely determined by the system noise (leading to $\mathcal{P}^*_t=\mathcal{L}^*_t$), in case 2, one can easily find that $\mathcal{P}^*_t\leq\mathcal{L}^*_t$, which reflects the additional influence of the parameter variation variance on $\mathcal{L}^*_t$.
\end{remark}
\paragraph{Numerical Demonstration.}
Experiments are conducted to illustrate the effectiveness of our algorithm. We assume the $d=2$ and introduce \emph{meta parameters} ($[4,2], [5.5,-1.5], [3,-1]$) to generate three groups of true parameters for 100 tasks by applying noise perturbation (i.e., $\mathcal{N}(0,0.5)$) to meta parameters. Then, for each task, 200 samples are generated using a linear regression model based on the sampled true parameter with noise perturbation (i.e., $\mathcal{N}(0,0.2)$), and then the 100 tasks are learned continually. In this case, there is no common global minimizer shared among all tasks, and the approximate common global minimizer is $[4,-\frac{1}{6}]$. We define two learning orders named sequential and random: at each stage, the former learns the tasks in the three meta groups sequentially while the latter learns a task random samples from 100 tasks.

We compare our algorithm with SGD in estimating the unknown approximate common global minimizer. Without regularization, the SGD Algorithm suffers from catastrophic forgetting (Figure \ref{figure-2} (a)), e.g., if the parameter of task $t+1$ is from another group, the estimator will directly overfit it, forgetting the current task $t$. Differently, our algorithm successfully finds the approximate common global minimizer in both sequential (Figure \ref{figure-2} (b)) and random (Figure \ref{figure-2} (c)) learning orders.

\section{Concluding Remarks}\label{conclu}
This paper establishes the theoretical foundation for continual learning, which is one of the central problems in current machine learning community. On the one hand, in the presence of a shared global minimizer, we demonstrate the global convergence of CL for a class of nonlinear regression models. This includes widely used models such as standard linear regression, logistic regression based classification, and the stochastic saturation model. On the other hand, we prove the global convergence of CL for general linear regressions when no shared global minimizer exists. To our best knowledge, our
work provides the first convergence results for continual learning under a non-i.i.d. and non-persistent excitation data assumptions.
Furthermore, without any excitation condition on the feature data, we establish the convergence rate for the forgetting and regret metrics for the first time, and the convergence rate of regret is potentially the best possible in the noise setting among all continual learners.
Several intriguing theoretical directions remain for further exploration, e.g., the convergence of nonlinear CL when no common global minimizer existing. We hope our work could provide a deep understanding of CL and facilitate the development of new, practical approaches.
\bibliography{example_paper}
\end{document}